\documentclass[11pt]{article}

\usepackage[final]{acl}

\usepackage{times}
\usepackage{latexsym}

\usepackage[T1]{fontenc}
\usepackage[utf8]{inputenc}

\usepackage{microtype}

\usepackage{inconsolata}

\usepackage{graphicx}

\usepackage{pifont}
\usepackage{hyperref}
\usepackage{url}
\usepackage{multirow}
\usepackage{booktabs}
\usepackage{algorithm}
\usepackage{amsmath,amssymb}
\usepackage[noend]{algpseudocode}
\usepackage{xspace}
\usepackage{booktabs}
\usepackage{threeparttable}
\usepackage{xcolor}
\usepackage{colortbl}
\usepackage{tabularx}
\usepackage{enumitem}
\usepackage{ragged2e} % for \RaggedRight
\usepackage{array}

\title{Stop When Enough: Adaptive Early-Stopping for Chain-of-Thought Reasoning}

\author{
  Renliang Sun$^{1}$\thanks{Equal Contributions.},
  Wei Cheng$^{2}$\footnotemark[1],
  Dawei Li$^{3}$\footnotemark[1],
  Haifeng Chen$^{2}$,
  Wei Wang$^{1}$ \\
  $^{1}$UCLA \quad $^{2}$NEC Labs America \quad $^{3}$Arizona State University \\
  \texttt{sunrenliang@ucla.edu, weiwang@cs.ucla.edu} \\
  \texttt{daweili5@asu.edu, \{weicheng, haifeng\}@nec-labs.com}
}

\begin{document}
\maketitle
\begin{abstract}
Chain-of-Thought (CoT) reasoning has driven recent gains of large language models (LLMs) on reasoning-intensive tasks by externalizing intermediate steps. 
% However, the widespread assumption that `more reasoning yields better performance' often leads to overthinking --- redundant steps that waste compute and can even hurt accuracy on simple problems.
However, excessive or redundant reasoning --- so-called overthinking --- can increase inference costs and lead LLMs toward incorrect conclusions.
In this paper, we present \textbf{REFRAIN} (\underline{REF}lective-\underline{R}edundancy for \underline{A}daptive \underline{IN}ference), a training-free framework that adaptively determines when to stop reasoning to mitigate overthinking.
REFRAIN integrates a two-stage stop discriminator to identify reflective yet redundant reasoning and a sliding-window Upper Confidence Bound (SW-UCB) multi-armed bandit controller to dynamically adjust stopping thresholds according to problem difficulty without supervision or fine-tuning.
Across four representative benchmarks and two model families, REFRAIN reduces token usage by 20-55\% while maintaining or improving accuracy compared to standard CoT prompting. Extensive ablation and robustness analyses demonstrate its stability across models, scorers, and prompt variations.
In summary, our findings highlight when-to-stop as a new and practical axis of test-time scaling --- enabling models to reason not just more, but just enough.
\end{abstract}
% \begin{abstract}
% Chain-of-Thought (CoT) prompting has been a critical technique for improving large language models (LLMs) on reasoning-intensive tasks. However, the assumption that longer CoTs always enhance accuracy often leads to overthinking-unnecessary reasoning not only lowers accuracy on simple questions but also inflates reasoning costs. In this work, we propose a dynamic framework that combines a two-stage stop discriminator for identifying reflective yet redundant reasoning steps with a sliding window upper confidence bound (SW-UCB) multi-armed bandit algorithm. This framework automatically adjusts the stop threshold to fit problems of different difficulty levels without supervision. Across four benchmarks and two LLMs, our approach consistently reduces token usage by 30-60\% while maintaining or even improving accuracy compared to vanilla CoT. Extensive analyses and diagnostic experiments further validate the robustness of our approach and its stability across different components, scorers, and prompting conditions. In summary, our results establish when-to-stop as a practical axis of test-time scaling, providing a simple yet effective way to mitigate overthinking in reasoning LLMs.
% \end{abstract}

\begin{figure*}[ht!]
    \centering
    \vspace{-0.2cm}
    \includegraphics[width=0.8\textwidth]{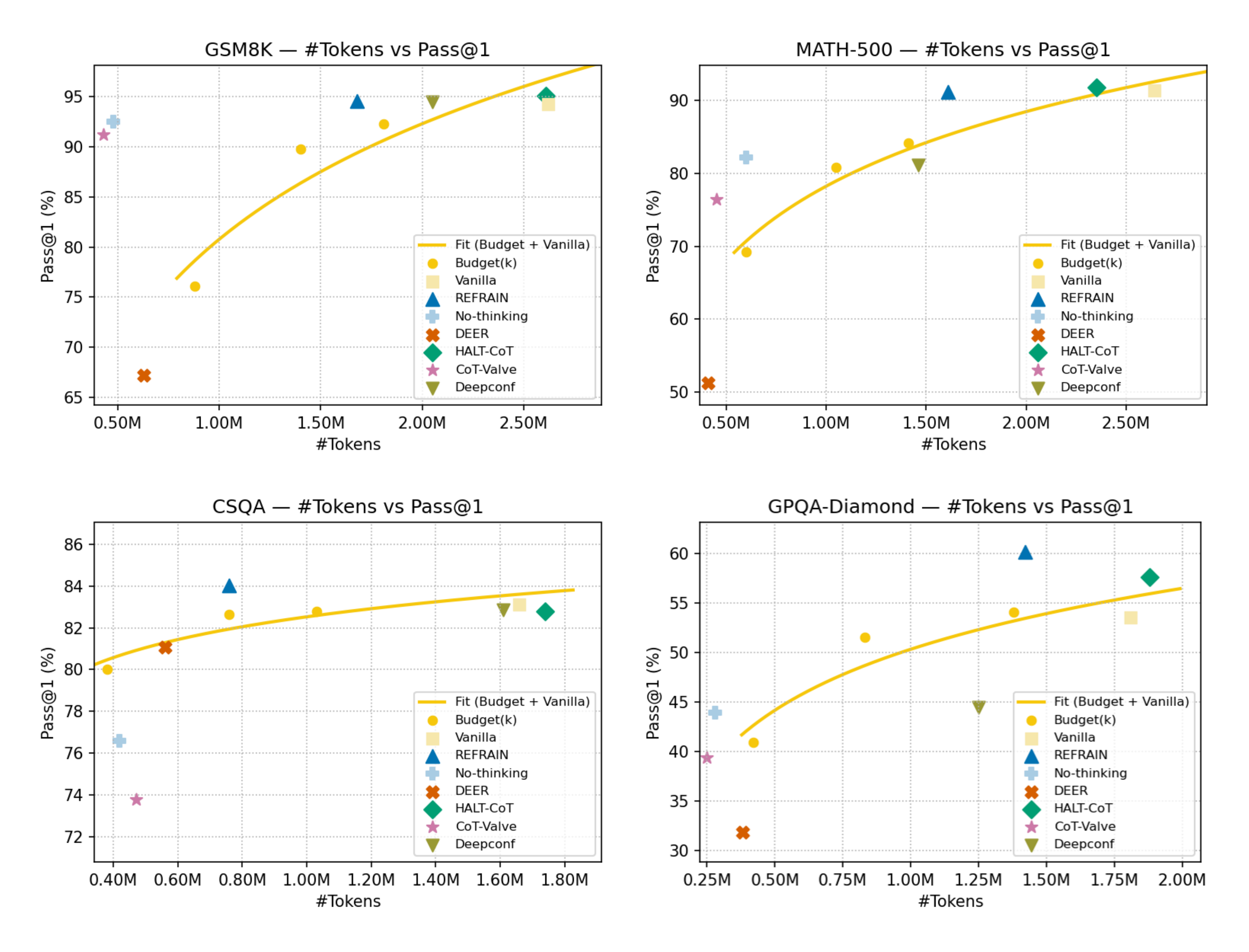}\vspace{-0.5cm}
    \caption{Test-time scaling with budgeted thinking: \#Tokens vs Pass@1 across four benchmarks using Qwen3-8B. We fit a log curve using budget points and vanilla. REFRAIN lies in the upper-left of the fitted curve, indicating a better accuracy-efficiency trade-off.}\vspace{-0.5cm}
    \label{fig:scaling}
\end{figure*}
\section{Introduction}
Large language models (LLMs) have recently achieved outstanding performance across a wide range of reasoning-intensive tasks, encompassing fields such as mathematics and common-sense reasoning \cite{tong2024can, guo2025deepseek, li2025system, yeo2025demystifying, yu2025causaleval}. 
A key driver of recent gains is that LLMs unfold a chain-of-thought (CoT) reasoning process before the final answer \cite{wei2022chain, xu2025softcot,zhao2025chain}. 
% However, while extended reasoning chains often correlate with improved accuracy, the field has overgeneralized this correlation into a rule, leading to a widespread but flawed belief: longer reasoning process always means better reasoning~\cite{jin2024impact, wu2025more, su2025between}.
However, a key limitation of these models is their tendency to \textit{overthink}—producing overly long reasoning trajectories filled with redundant or irrelevant steps~\cite{chen2024not,cui2025stepwise}. Such behavior not only increases inference overhead but can also steer the model toward incorrect conclusions~\cite{jiang2025drp,sui2025stop}.

% A key technique that has driven these improvements is the use of \textit{Chain-of-Thought} (CoT) prompting, where the model is encouraged to generate explicit intermediate reasoning steps before producing the final answer \cite{wei2022chain, xu2025softcot}. However, while extended reasoning chains often correlate with improved accuracy, the field has overgeneralized this correlation into a rule, leading to a widespread but flawed belief: longer thinking always means better thinking~\cite{jin2024impact, wu2025more, su2025between}. 
% \dawei{The logic here is not correct to me; people actually know that longer reasoning does not necessarily lead to more accurate answers. The problem is that overthinking is an intrinsic problem of LRMs.}

% This overemphasis on verbose reasoning has created a new bottleneck \cite{cuadron2025danger, pu2025thoughtterminator}. Existing LLMs are increasingly falling into the trap of `overthinking', wasting computational resources on redundant reasoning that fails to improve accuracy~\cite{wu2025when, huang2025mitigating}. Recent works such as HALT-CoT \cite{laaouach2025halt} attempt to address this by introducing confidence or entropy-based early stopping signals, but these methods rely on manually designed heuristics for each task. Other methods like CoT-Valve \cite{ma2025cot} require additional computational resources to learn short-thinking behaviors.

Recent works such as HALT-CoT \cite{laaouach2025halt} attempt to address this by introducing confidence or entropy-based early stopping signals, but these methods rely on manually designed heuristics for each task. Other methods like CoT-Valve \cite{ma2025cot} and Deepconf~\cite{fu2025deep} require additional computational resources to learn short-thinking behaviors or search for the best stopping threshold. While they achieve promising results, their reliance on manual heuristics and additional computational resources makes them difficult to generalize across diverse tasks and impractical in low-resource reasoning scenarios.

% or supervised thresholds that fail to adapt to varying task complexities\dawei{why?}. 
% In practice, such approaches are brittle: they either stop too early on difficult problems or waste computation on trivial ones. \dawei{This is a general problem if early stop doesn't work well.}

% We argue that the challenge is not only stopping earlier, but stopping optimally-adapting the reasoning depth dynamically to each task's difficulty.
% This requires a framework that (1) detects when reasoning stops to yield new information, (2) adapts its behavior online without supervision or retraining, and (3) maintains or improves accuracy while saving inference cost.

We argue that the core challenge lies not merely in stopping earlier, but in automatically and efficiently determining an optimal threshold that dynamically adapts the reasoning depth to each task’s difficulty. This calls for a framework that \ding{182} automatically detects when further reasoning becomes redundant, \ding{183} adapts its stopping behavior online without external supervision or retraining, and \ding{184} maintains or even improves accuracy while reducing inference cost.

To this end, we introduce \textbf{REFRAIN} (\underline{REF}lective-\underline{R}edundancy for \underline{A}daptive \underline{IN}ference), a training-free, dynamic framework that transforms `when-to-stop' into a new axis of test-time scaling for reasoning LLMs. REFRAIN integrates two synergistic ideas:

\begin{itemize}[leftmargin=0.5pt, itemindent=1.2em, itemsep=0.0em, 
topsep=0.0em,
parsep=0.2em]
\item{Reflective redundancy detection} --- a discriminator that identifies when reasoning transitions from reflective self-correction to redundant repetition, based on semantic similarity and trigger cues.
\item{Adaptive thresholding via Sliding-Window Upper Confidence Bound (SW-UCB)} --- a multi-armed bandit controller that continuously balances exploration (longer reasoning when uncertain) and exploitation (early stop when confident), enabling task-specific efficiency.
\end{itemize}

Across four representative benchmarks and two model families, REFRAIN achieves 20-55\% fewer tokens while maintaining or exceeding vanilla CoT accuracy, outperforming prior stopping methods and even matching fine-tuned baselines. As shown in Figure \ref{fig:scaling}, REFRAIN consistently shifts the accuracy–efficiency frontier upward and leftward relative to vanilla CoT.
Thus, REFRAIN scales reasoning not by thinking more, but by allocating thought where it matters.
% Our results demonstrate that adaptive stopping is not just an efficiency trick --- it is a scalable reasoning principle. By converting `how much to think' into an actionable control variable, REFRAIN reframes inference-time optimization as budgeted intelligence: learning not only to reason better, but to reason just enough.

% ~\dawei{too general, we should mention more specific advantages of our method}.

In summary, this work makes the following key contributions:

\begin{itemize}[leftmargin=0.5pt, itemindent=1.2em, itemsep=0.0em, 
topsep=0.0em,
parsep=0.2em]
    \item We identify reflective redundancy as the underlying signal behind overthinking and make it actionable through a two-stage stop discriminator.
    \item We propose a training-free adaptive thresholding algorithm based on SW-UCB bandits that dynamically balances exploration and exploitation during reasoning.
    \item We show consistent accuracy-efficiency gains across diverse reasoning tasks, establishing when-to-stop as a practical and generalizable dimension of test-time scaling.
\end{itemize}

\section{Related Work}
\subsection{Overthinking in LLMs}

Recent work characterizes overthinking in reasoning LLMs as \textit{reasoning beyond what is needed to solve a problem}. \citet{chen2024not} show it is widespread: redundant steps add little to correctness or diversity, yet waste computation on simple problems. \citet{sui2025stop} define it as verbose, redundant outputs that induce substantial overhead.

Beyond definitional studies, several works have empirically analyzed how overthinking manifests and affects reasoning performance. \citet{chiang2024over} showed that models often conduct redundant reasoning even for simple problems, sometimes leading to errors. \citet{fan2025missing} show that with missing premises, models generate lengthy yet unhelpful chains instead of stopping. \citet{gema2025inverse} report an inverse scaling effect: longer reasoning can reduce accuracy. \citet{cuadron2025danger} analyzed forms that can impair accuracy, such as analysis paralysis and rogue actions. Collectively, these studies show overthinking is inefficient and can degrade reliability, motivating methods that dynamically regulate reasoning length.

% Some research work has proposed metrics and evaluation methods to quantify their degree. For instance, Fan et al. (2025) proposed the `rejection rate' that the frequency with which the model appropriately declines to answer unanswerable queries. Cuadron et al. (2025) categorized manifestations like analysis paralysis and rogue actions and designed a scoring system to quantify overthinking behaviors in agentic tasks.

\subsection{Overthinking Mitigation}
\label{sec:reduce overthinking}
Existing efforts to mitigate overthinking generally fall into two categories: (1) post-training methods that teach models to shorten reasoning without harming accuracy, and (2) inference-time strategies that adaptively decide when to stop reasoning.

Representative post-training methods include CoT-Valve \cite{ma2025cot}, which fine-tunes LLMs to control reasoning length by identifying parameter-space `valves' that adjust the verbosity of generated thoughts; AALC \cite{li2025aalc}, which uses an RL reward to balance correctness and conciseness; and SmartThinker \cite{he2025smartthinker}, which combines SFT on short-form data with RL that allocates tokens to critical steps.

%AALC \cite{li2025aalc}, which introduces a reinforcement learning reward balancing correctness and conciseness by penalizing verbose reasoning only after achieving high-accuracy goals; and SmartThinker \cite{he2025smartthinker}, which combines supervised fine-tuning on short-form reasoning data with reinforcement learning that allocates more tokens to critical steps.

% One research direction involves using post-training on existing models to enable them to learn how to shorten reasoning length without sacrificing accuracy. Recent work includes CoT-Valve, AALC, SmartThinker, and many others. CoT-Valve \cite{ma2025cot} controls the length of its chain-of-thought reasoning by fine-tuning LLMs. By identifying specific parameter space directions called `valves', the model can generate shorter or longer reasoning chains as needed. AALC \cite{li2025aalc} introduces a reinforcement learning reward mechanism that balances correctness and conciseness by penalizing verbose reasoning only after achieving high-accuracy goals. SmartThinker \cite{he2025smartthinker} achieves reasoning compression and preservation through step-size control. First, the LLM is supervised-fine-tuned using `short-form' reasoning data. Then, a customized reinforcement learning process allocates more tokens to critical reasoning steps.

In contrast, inference-time approaches avoid retraining and focus on adaptive stopping --- deciding when the model has reasoned enough. They seek reliable stopping signals at inference.
Methods such as ESC \cite{li2024escape} and s1 \cite{muennighoff2025s1} regulate reasoning by monitoring the stability of the answer distribution. ESC dynamically stops sampling once the predicted answer converges, whereas s1 inserts wait tokens to postpone termination until enough evidence is gathered.

Other methods use internal confidence or entropy for early stopping. HALT-CoT \cite{laaouach2025halt} stops reasoning when prediction entropy drops below a threshold, while DEER \cite{yang2025dynamic} and DeepConf \cite{fu2025deep} exploit confidence signals to prune redundant reasoning paths, enhancing efficiency and accuracy. AlphaOne \cite{zhang2025alphaone} and No-thinking \cite{ma2025reasoning} use deterministic or skip-based mechanisms to stop once enough information is available.

Overall, existing methods either rely on supervised fine-tuning or handcrafted confidence signals. However, few provide a unified unsupervised framework that adaptively balances reasoning depth and efficiency across tasks, which is a challenge that our work aims to address.

\section{Methodology}

\subsection{Definition}

To formally describe our adaptive stopping mechanism, we first define the reasoning process of an LLM.
Given a question $Q$, the LLM model $p_{\theta}(\cdot)$ produces a sequence of reasoning steps:

\begin{equation}
    S = (s_1, \ldots , s_N)
    % , \qquad s_n \in \mathcal{X}
\end{equation}

\noindent followed by a final answer $y$. In practice, reasoning steps are segmented from the decoded text using blank-line delimiters. Our goal is to identify the optimal stopping point—halting generation once further reasoning becomes redundant—and produce a concise, well-formed final answer.

\begin{figure*}[ht!]\vspace{-0.45cm}
    \centering
    \includegraphics[width=0.95\textwidth]{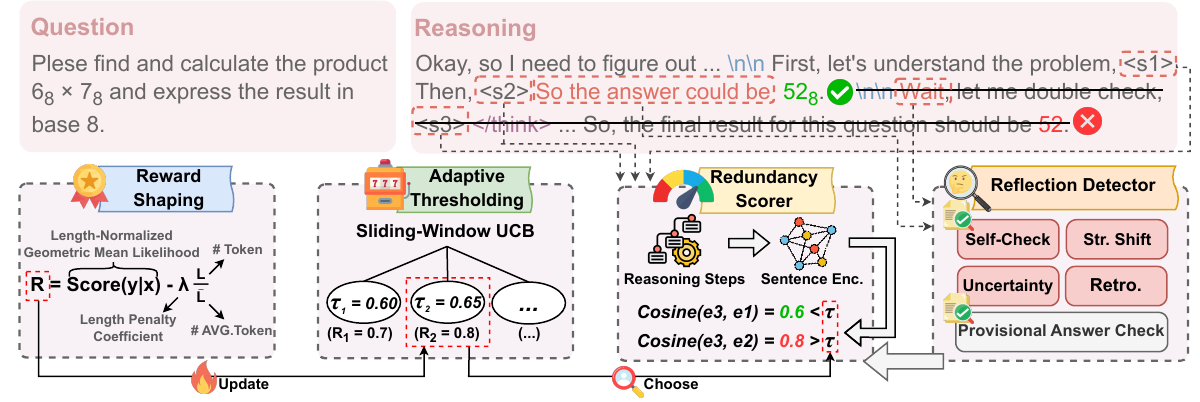}
    \caption{Overview of the proposed REFRAIN method.}
    \label{fig:method}\vspace{-0.3cm}
\end{figure*}

\subsection{Two-Stage Stopping Discriminator}
\label{subsec:discriminator}

We propose a discriminator \(D\) that triggers at step \(n\) only when (i) the step is \emph{reflective} and (ii) its content is \emph{semantically redundant} relative to the previous steps. The pseudocode for implementing discriminator $D$ is shown in Algorithm \ref{alg:stop}.

\paragraph{Reflection detector.}
We first detect reflective reasoning cues. 
Following prior analyses of reflective or self-corrective cues in CoT reasoning \cite{ma2024large, yang2025speculative, huang2025efficient, qiao2025concise}, we define reflection operationally through a consolidated vocabulary derived from these studies.
Concretely, we adopt the category structure reported in existing studies --- self check, strategy shift, uncertainty expression, and retrospective revisions --- and merge the trigger terms into four sets \(V_{\text{check}}, V_{\text{shift}}, V_{\text{uncert}}, V_{\text{retro}}\). The full term lists used in our experiments are documented in Appendix A to facilitate replication. Let \(V=V_{\text{check}}\cup V_{\text{shift}}\cup V_{\text{uncert}}\cup V_{\text{retro}}\) denote the complete reflection trigger vocabulary. We flag step \(s_n\) as reflective if any \(v\in V\) occurs:

\begin{equation}
    r_n=\mathbb{I}\!\left(\exists\,v\in V\ \text{s.t.}\ v\subseteq s_n\right)
\end{equation}

\noindent where $\mathbb{I}(\cdot)$ is the indicator function which returns $1$ if the condition inside holds and $0$ otherwise. To prevent premature triggers early in the trace, we additionally require the history to include a provisional answer cue 
$c$ (e.g., `answer is/should be'), defined as:
\vspace{-0.15cm}
\begin{equation}
h_{n-1} = \mathbb{I}\left(\exists j < n \text{ s.t. } c \subseteq s_j\right)
\label{eq:ensure_answer}
\end{equation}

% Equation \ref{eq:ensure_answer} ensures that reflection is only considered valid when the reasoning history already contains a provisional answer, preventing premature triggers during exploration.

\paragraph{Semantic redundancy scorer.}
We embed each step with a sentence encoder \(f(\cdot)\) (we use \texttt{all-MiniLM-L6-v2} in practice) and compute the maximum cosine similarity to the previous steps:
\vspace{-0.16cm}
\begin{equation}
    e_n=f(s_n),\qquad \phi_n=\max_{1\le j<n}\cos\!\big(e_n,e_j\big).
\end{equation}
Given the threshold \(\tau\in[0,1]\), the stop rule is:
\begin{equation}
\begin{split}
    \textsc{Stop}(Q,s_{1:n}) \;=\;& \mathbb{I}(h_{n-1}=1)\cdot\mathbb{I}(r_n=1)\\
    &\cdot\mathbb{I}(\phi_n\ge\tau).
\end{split}
\end{equation}

Once triggered, we halt further reasoning and proceed to the final answer generation stage.

% \textbf{Semantic Redundancy Scorer}

% For the semantic redundancy scorer, we use a sentence vector encoder $f(\cdot)$ to embed each reasoning step. Then, we defined the maximum similarity $\phi_n$ between $s_n$ and all previous reasoning steps:

% \begin{equation}
% e_n = f(s_n) \qquad \phi_n = \max_{1 \le j < n} \cos(e_n, e_j)
% \end{equation}

% In practice, we employ the `all-MiniLM-L6-v2' model as the sentence vector decoder $f(\cdot)$. Let $\tau \in [0,1]$ be a redundancy threshold, the rule to halt thinking is:

% \begin{equation}
% Stop(Q, s_{1:n}) = \mathbb{I}(h_{n-1} = 1) \cdot \mathbb{I}(r_n = 1) \cdot \mathbb{I}(\phi_n \ge \tau)
% \end{equation}

% Once triggered, we stop further thinking and jump to the answer generation and extraction stage.

\begin{algorithm}[t]\vspace{-0.1cm}
\caption{Early-Stop via Reflective Redundancy}
\label{alg:stop}
\small
\begin{algorithmic}[1]
\Require Question \(Q\); model \(p_\theta\); encoder \(f\); trigger sets \(V\), cue \(c\); threshold \(\tau\).
\State Initialize step list \(S\gets[~]\), embeddings \(E\gets[~]\), flags \(h\gets 0\).
\For{\(n=1,2,\ldots\)}
    \State Generate next step \(s_n\) (until blank-line delimiter).
    \State \(S\gets S\mathbin{\|} s_n\); \quad \(e_n\gets f(s_n)\); \quad \(E\gets E\mathbin{\|} e_n\).
    \State \(r_n\gets \mathbb{I}(\exists v\in V: v\subseteq s_n)\).
    \State \(h\gets h \lor \mathbb{I}(\exists j<n: c\subseteq s_j)\).
    \State \(\phi_n\gets \max_{1\le j<n}\cos(e_n, e_j)\) \textbf{ (if \(n>1\), else \(0\)) }.
    \If{\(h=1\) \textbf{ and } \(r_n=1\) \textbf{ and } \(\phi_n\ge\tau\)}
        \State \textbf{break} \Comment{stop thinking}
    \EndIf
\EndFor
\State \Return stop trace \(S\)
\end{algorithmic}\vspace{-0.16cm}
\end{algorithm}
\subsection{Answer-Only Likelihood Scoring}\vspace{-0.1cm}

After stopping the thinking process, we elicit an answer using a forced-closure prompt: $\texttt{Final Answer: \textbackslash boxed\{}\cdots\texttt{\}}$. 
To mitigate stylistic variance in reasoning traces, we evaluate only the answer token sequence 
\(y\) within the boxed region.
% To reduce the noise of stylistic differences in reasoning, we score only the answer content \(y\) inside the box. 
Let \(x\) denote the answer prefix context, we use the length-normalized geometric mean likelihood:
\vspace{-0.16cm}
\begin{equation}
\mathbf{Score}(y \mid x) = \exp\left(\frac{1}{|y|}\sum_{i=1}^{|y|}\log p_{\theta}(y_i | x, y_{<i})\right)
\end{equation}
% \dawei{Use for what? It could be confusing why we need Score(y given x) here.}
% The design of only scoring the final answer directly aligns rewards with answer quality, mitigating noise from differences in reasoning expressions and facilitating the feedback of reward signals to the stop policy.

\subsection{Adaptive Thresholding via Sliding-Window UCB}

Due to the varying complexity of reasoning across different tasks, employing a uniform fixed threshold $\tau$ is not the optimal solution, and manually defining an optimal threshold for each task is also extremely challenging. Therefore, we adaptively select $\tau$ from a discrete set $\mathcal{T}$ with a sliding-window UCB (SW-UCB) bandit \cite{garivier2008upper}. The pseudocode is shown  in Algorithm \ref{alg:swucb}. For each arm $t \in \mathcal{T}$, we maintain a window of the most recent $W$ rewards ${R^{(t)}}$ with running mean ${\bar{R}^{(t)}}$. Let \(n^{(t)}\) be the number of pulls of arm \(t\) in the window, and define the effective time
\begin{equation}
    t_{\mathrm{eff}}=\min\!\big(k,\; W\cdot|\mathcal{T}|\big),
\end{equation}
where \(k\) is the round index. We then select
\begin{equation}
    t_k=\arg\max_{t\in\mathcal{T}}~ \bar{R}^{(t)} + C\sqrt{\frac{2\log t_{\mathrm{eff}}}{\,n^{(t)}\,}},
\end{equation}
with each arm explored once for initialization to ensure unbiased reward estimation.

\begin{algorithm}[t]
\caption{SW-UCB for Adaptive \(\tau\) Selection}
\label{alg:swucb}
\small
\begin{algorithmic}[1]
\Require Candidate set \(\mathcal{T}\); window size \(W\); exploration constant \(C>0\).
\State Initialize buffers \(\{\mathcal{B}^{(t)}\}_{t\in\mathcal{T}}\leftarrow\) empty deques of capacity \(W\).
\For{\(k=1,2,\ldots\)}
    \State \textbf{Cold-start:} if \(\exists\, t\in\mathcal{T}\) with \(|\mathcal{B}^{(t)}|=0\), set \(t_k\gets t\) (arbitrary such \(t\)).
    \State \textbf{Otherwise:}
        \State \(\displaystyle t_{\mathrm{eff}}\gets \min\big(k,\, W\cdot|\mathcal{T}|\big)\).
        \ForAll{\(t\in\mathcal{T}\)}
            \State \(\bar{R}^{(t)}\gets \mathrm{mean}(\mathcal{B}^{(t)})\); \quad \(n^{(t)}\gets |\mathcal{B}^{(t)}|\).
            \State \(\mathrm{UCB}^{(t)}\gets \bar{R}^{(t)} + C\sqrt{2\log(t_{\mathrm{eff}})/\max(1,n^{(t)})}\).
        \EndFor
        \State \(t_k\gets \arg\max_{t\in\mathcal{T}} \mathrm{UCB}^{(t)}\).
    \State Run Alg.~\ref{alg:stop} with \(\tau=t_k\); obtain stopped trace and final answer \(y_k\).
    \State Compute reward \(R_k\) (Eq.~\ref{eq:reward}); push \(R_k\) into \(\mathcal{B}^{(t_k)}\) (evict oldest if full).
\EndFor
\end{algorithmic}
\end{algorithm}

\subsection{Reward Shaping}

To guide the adaptive selection process, we design a reward function that jointly accounts for answer quality and reasoning efficiency. Let \(L\) be the total output token count (thinking + answer), and \(\bar{L}\) its running mean. We define
\begin{equation}
\label{eq:reward}
    R \;=\; \mathrm{Score}(y\mid x) \;-\; \lambda\,\frac{L}{\bar{L}},
\end{equation}
with \(\lambda>0\). For the very first sample where \(\bar{L}\) is undefined, we use a linear cold start penalty \(0.0001\cdot L\). This reward formulation encourages accurate yet concise reasoning, guiding the bandit to prefer earlier stopping (i.e., larger $\tau$) when answer confidence is high. 
We deliberately avoid process evaluators to keep REFRAIN training-free and single-forward at test time. When process fidelity is paramount, more expensive rewards can be plugged into Algorithm \ref{alg:swucb} without changing the current early-stop mechanism.
% \dawei{We deliberately avoid process evaluators (e.g., PRM/RM) to keep REFRAIN training-free and single-forward at test time. When process fidelity is paramount, more expensive rewards can be plugged into Algorithm \ref{alg:swucb} without altering the early-stop mechanism.}

\subsection{Online-Adaptive Thresholding}

Figure \ref{fig:method} provides an overview of our method, REFRAIN (REFlective-Redundancy for Adaptive INference). REFRAIN operates entirely at test time, requiring no fine-tuning or validation data. The sliding-window UCB controller dynamically adjusts the threshold $\tau$ based on recent reward feedback, allowing the model to balance accuracy and efficiency across varying task difficulties and to autonomously determine when to stop reasoning.

\section{Experimental Settings}

\subsection{Models}

We evaluate our approach using two representative reasoning-oriented LLMs:

\noindent \textbf{Qwen3-8B.} Qwen3-8B \cite{yang2025qwen3} is a dense decoder-only model with approximately 8 billion parameters.

\noindent \textbf{gpt-oss-20B.} gpt-oss-20B \cite{agarwal2025gpt} is a mixture-of-experts (MoE) model with 21 billion total parameters and 3.6 billion active parameters per forward pass.

More details of the implementation can be found in Appendix E.

% To achieve optimal performance, we used the officially recommended decoding settings. For Qwen3-8B, we enabled thinking mode and set Temperature=0.6, TopP=0.95, TopK=20. For gpt-oss-20B, we used BF16 precision and set Temperature=1.0, TopP=1.0, TopK=50.

\subsection{Benchmarks}

We conduct experiments on four benchmarks that cover mathematical problems, common-sense reasoning problems, and stem problems:

\noindent \textbf{GSM8K.} GSM8K \cite{cobbe2021training} consists of 1,319 grade-school math problems requiring multi-step arithmetic reasoning. 
% We evaluate on the official test set.

\noindent \textbf{MATH-500.} MATH-500 \cite{hendrycks2021measuring} consists of 500 challenging math problems selected from the MATH benchmark. % covering subfields such as algebra, number theory, and geometry.

\noindent \textbf{CommonsenseQA (CSQA).} CSQA \cite{talmor2018commonsenseqa} is a multiple-choice common-sense reasoning dataset with 1,221 questions. Since the official test set does not provide labels, we report results on the validation set.

\noindent \textbf{GPQA-Diamond.} GPQA-Diamond \cite{rein2024gpqa} is the most challenging subset of the GPQA benchmark, comprising 198 graduate-level interdisciplinary questions in a multiple-choice format.

\subsection{Evaluation}

We evaluate model performance in terms of accuracy and efficiency using two metrics:

\noindent \textbf{Pass@1.} For GSM8K and MATH-500, we extract the final boxed expression or numerical answer from the model's output and apply strict string matching to ground truth. For CSQA and GPQA-Diamond, we evaluate multiple-choice accuracy by comparing the predicted option (A–E) with the gold label.

\noindent \textbf{\#Tokens.} We measure the number of tokens generated (reasoning steps + final answer). At comparable accuracy levels, lower token usage indicates more efficient reasoning and reduced overthinking.

% The first one is \textbf{Accuracy}, which measures how well the model performs on these datasets. For GSM8K and MATH-500, we parse the model's output for answers into final numerical values or expressions and perform strict regular expression matching. For CSQA and GPQA-diamond, we compare the consistency between the options provided by the model (A/B/C/D/E) and the standard answers. The second metric is the number of generated \textbf{Tokens}. At comparable accuracy levels, fewer generated tokens indicate the model's stronger ability to avoid overthinking.
\section{Results}

In this section, we first present the overall performance of REFRAIN across benchmarks and models (§5.1), followed by ablation and robustness studies (§5.2-5.6) that examine the contribution of individual components and generalization across settings.

% ---- 放在导言区（或表格前）----
\makeatletter
\providecommand{\textsubscript}[1]{$_{\textstyle #1}$}
\makeatother
\newcommand{\chgpass}[1]{\textsubscript{\smash{\scriptsize #1}}}
\newcommand{\chgtok}[1]{\textsubscript{\smash{\scriptsize #1}}}
% ---------------------------------

\subsection{Main Results: Accuracy-Efficiency Trade-off}

\begin{table*}[ht!]
\vspace{-0.5cm}
\centering
\small
\resizebox{\textwidth}{!}{
\begin{threeparttable}
\setlength{\tabcolsep}{6pt}
\begin{tabular}{@{}l|cc|cc|cc|cc@{}}
\toprule
\multirow{2}{*}{\textbf{Qwen3-8B}} 
& \multicolumn{2}{c|}{\textbf{GSM8K}} 
& \multicolumn{2}{c|}{\textbf{MATH-500}} 
& \multicolumn{2}{c|}{\textbf{CSQA}} 
& \multicolumn{2}{c}{\textbf{GPQA-Diamond}} \\
\cmidrule(lr){2-3} \cmidrule(lr){4-5} \cmidrule(lr){6-7} \cmidrule(lr){8-9}
& \textbf{Pass@1} & \textbf{\#Tokens} 
& \textbf{Pass@1} & \textbf{\#Tokens} 
& \textbf{Pass@1} & \textbf{\#Tokens}
& \textbf{Pass@1} & \textbf{\#Tokens} \\
\midrule
Vanilla     & 94.24 & 2.62M & 91.40 & 2.64M & 83.13 & 1.66M & 53.54 & 1.81M \\
No-thinking & 92.49 & 0.48M & 82.20 & 0.60M & 76.58 & \textbf{0.42M} & 43.94 & 0.28M \\
DEER        & 67.25 & 0.63M & 51.20 & \textbf{0.41M} & 81.08 & 0.56M & 31.82 & 0.38M \\
HALT-CoT    & \textbf{95.07} & 2.61M & \textbf{91.80} & 2.35M & 82.80 & 1.74M & 57.58 & 1.88M \\
\midrule
CoT-Valve   & 91.21 & \textbf{0.43M} & 76.40 & 0.45M & 73.79 & 0.47M & 39.39 & \textbf{0.25M} \\
Deepconf    & 94.47 & 2.05M & 81.20 & 1.46M & 82.87 & 1.61M & 44.44 & 1.25M \\
\midrule
REFRAIN     & 94.54\chgpass{+0.30} & 1.68M\chgtok{-35.9\%} & 91.20\chgpass{-0.20} & 1.61M\chgtok{-39.0\%} & \textbf{84.03}\chgpass{+0.90} & 0.76M\chgtok{-54.2\%} & \textbf{60.10}\chgpass{+6.56} & 1.42M\chgtok{-21.5\%} \\
\bottomrule
\end{tabular}
\vspace{0.1cm}
\begin{tabular}{@{}l|cc|cc|cc|cc@{}}
\toprule
\multirow{2}{*}{\textbf{gpt-oss-20B}} 
& \multicolumn{2}{c|}{\textbf{GSM8K}} 
& \multicolumn{2}{c|}{\textbf{MATH-500}} 
& \multicolumn{2}{c|}{\textbf{CSQA}} 
& \multicolumn{2}{c}{\textbf{GPQA-Diamond}} \\
\cmidrule(lr){2-3} \cmidrule(lr){4-5} \cmidrule(lr){6-7} \cmidrule(lr){8-9}
& \textbf{Pass@1} & \textbf{\#Tokens} 
& \textbf{Pass@1} & \textbf{\#Tokens} 
& \textbf{Pass@1} & \textbf{\#Tokens}
& \textbf{Pass@1} & \textbf{\#Tokens} \\
\midrule
Vanilla     & 91.66 & 0.76M & 80.80 & 1.07M & \textbf{82.96} & 0.80M & 34.85 & 0.93M \\
No-thinking & 90.83 & \textbf{0.38M} & 74.20 & 0.43M & 76.33 & \textbf{0.40M} & 30.30 & 0.23M \\
DEER        & 93.63 & 0.63M & 83.20 & 0.76M & 70.52 & 1.14M & 15.66 & 0.67M \\
HALT-CoT    & 92.65 & 0.50M & 79.80 & 0.66M & 77.15 & 0.60M & 40.40 & 0.50M \\
\midrule
CoT-Valve   & 89.08 & 0.43M & 62.20 & \textbf{0.32M} & 71.42 & 0.49M & 15.15 & \textbf{0.14M} \\
Deepconf    & 93.33 & 0.58M & 83.00 & 0.60M & 81.00 & 0.75M & 25.76 & 0.73M \\
\midrule
REFRAIN     & \textbf{94.39}\chgpass{+2.73} & 0.42M\chgtok{-44.7\%} & \textbf{84.20}\chgpass{+3.40} & 0.69M\chgtok{-35.5\%} & 81.74\chgpass{-1.22} & 0.45M\chgtok{-43.8\%} & \textbf{41.92}\chgpass{+7.07} & 0.62M\chgtok{-33.3\%} \\
\bottomrule
\end{tabular}
\end{threeparttable}}

\caption{Overall accuracy–efficiency results across four benchmarks (Pass@1$\uparrow$ / \#Tokens$\downarrow$). Vanilla means we use default generation settings with no early stopping. CoT-Valve requires fine-tuning, while Deepconf needs additional CoT generation. \textbf{Bold} marks the best value per column. For REFRAIN, subscripts denote its change vs.\ Vanilla: Pass@1 uses $+\!/-$ absolute points, \#Tokens shows percentage reduction.}
\vspace{-0.2cm}
\label{tab:main results}
\end{table*}

We evaluated REFRAIN against baselines on four benchmarks. Our baselines are No-thinking \cite{ma2025reasoning}, DEER \cite{yang2025dynamic}, HALT-CoT \cite{laaouach2025halt}, CoT-Valve \cite{ma2025cot}, and Deepconf \cite{fu2025deep}. We introduced these methods in Section \ref{sec:reduce overthinking}, so we briefly summarized them here. For No-thinking, the reasoning trace is replaced by a fixed phrase `\texttt{Okay, I think I have finished thinking.}'. For DEER and HALT-CoT, we have reproduced the methods described in the paper. For CoT-Valve, we have fine-tuned Qwen3 and gpt-oss using the short thinking process detailed in the paper. For Deepconf, we have reproduced the method and adapt it to our single-trace setting.
Notably, all methods except CoT-Valve and Deepconf require no additional fine-tuning or CoT generation.

Table \ref{tab:main results} compares REFRAIN with the baselines across all benchmarks. Overall, REFRAIN attains vanilla-level or higher Pass@1 while consuming 20-55\% fewer tokens, delivering the best accuracy-efficiency balance among training-free methods and competitive against fine-tuned baselines.

% For Qwen3-8B, REFRAIN achieves similar results to Vanilla in Pass@1 while cutting tokens by \textasciitilde40\% on GSM8K and MATH-500. On CSQA, REFRAIN slightly surpasses Vanilla in Pass@1 with a \textasciitilde55\% token reduction. On the harder benchmark GPQA-Diamond, REFRAIN improves Pass@1 materially with \textasciitilde20\% fewer tokens.

For Qwen3-8B, REFRAIN matches the vanilla accuracy on GSM8K and MATH-500 while cutting token usage by approximately 40\%. It slightly surpasses vanilla on CSQA with a 55\% reduction in tokens, and yields notable improvements on the more challenging GPQA-Diamond benchmark with around 20\% fewer tokens.

Similar trends are observed with gpt-oss-20B.
REFRAIN shows obvious improvements in Pass@1 compared to Vanilla on GSM8K, MATH-500, and GPQA-Diamond, while maintaining a comparable Pass@1 on CSQA. REFRAIN also achieved token savings of \textasciitilde45\%, \textasciitilde35\%, \textasciitilde45\%, and \textasciitilde35\% respectively across the four benchmarks. These consistent cross-model trends suggest REFRAIN’s benefits are architecture-agnostic.

We also compare REFRAIN with other representative baselines. No-thinking uses the fewest tokens but lowers Pass@1, especially on the higher-difficulty benchmarks MATH-500 and GPQA-Diamond. DEER and CoT-Valve often save tokens, but Pass@1 significantly drops on some benchmarks. For example, when using Qwen3, DEER performs noticeably worse than Vanilla on GSM8K and MATH-500. When using gpt-oss, CoT-Valve performs poorly on GPQA-Diamond. HALT-CoT maintains Pass@1 but saves little tokens. Deepconf shows competitive performance on some benchmarks such as GSM8K and MATH-500, but is inconsistent in others, such as the GPQA-Diamond benchmark when using gpt-oss.

Overall, REFRAIN systematically curbs overthinking by detecting reflective yet redundant steps and stopping early when additional reasoning is unlikely to help. This yields substantial token reductions at equal or higher Pass@1 across tasks and backbones, positioning when-to-stop as a practical axis of test-time scaling that complements longer chains and larger models.

\subsection{Completeness of Trigger Vocabulary and Categories}
\label{sec:triggers}

To assess the adequacy of our four predefined categories of reflective triggers,
%: self-check, strategy shifts, uncertainty, and retrospective revisions
we conducted three controlled experiments on the MATH-500 dataset using Qwen3-8B. All other configurations remained consistent with the main experiment.

\noindent \textbf{Leave-One-Category-Out.} We ablated one trigger category at a time while keeping the others unchanged. 
% This allows us to examine whether any single category is indispensable for achieving strong performance.

% \textbf{Leave One Category Out}: While keeping the other three categories unchanged, remove one trigger word category at a time to observe how accuracy and total token usage change.

\noindent \textbf{In-Category Vocabulary Expansion (In-cat Expansion).} We augmented each category with natural synonyms and semantically similar expressions. This tests whether enlarging the lexical coverage within categories yields additional benefits.

% \textbf{In-Category Vocabulary Expansion (Vocab Expansion)}: Without changing the categories, supplement each category with natural synonyms/similar expressions to examine whether they bring additional benefits and costs.

\noindent \textbf{New Category Addition (New Category).} We introduced an additional trigger category with no semantic overlap with the existing categories to evaluate whether additional functional categories can further improve performance.

% \textbf{Add a New Category (New Category)}: Introduce a category that does not overlap semantically with the first four categories. Evaluate whether it can further enhance performance or reduce thinking costs.

For the second and third experiments, we used GPT-5 to generate candidate synonyms and new category terms, which were subsequently filtered to ensure semantic validity. The complete vocabulary is listed in Table \ref{tab:reflection-lexicon}.

\begin{table}[ht]
\vspace{-0.2cm}
\centering
\small
\begin{threeparttable}
\setlength{\tabcolsep}{8pt}
\resizebox{0.4\textwidth}{!}{
\begin{tabular}{@{}l|cc@{}}
\toprule
\textbf{Variants} & \textbf{Pass@1}$\uparrow$ & \textbf{\#Tokens}$\downarrow$ \\
\midrule
Vanilla & 91.40 & 2.64M \\
REFRAIN & 91.20 & 1.61M \\
\midrule
\addlinespace[2pt]
\multicolumn{3}{c}{\textit{Shorten the Vocabulary}} \\
\midrule
% \cmidrule(lr){1-3}
w/o self-check & 90.20 & 1.77M \\
w/o strategy-shift & 90.20 & 1.68M \\
w/o uncertainty & 90.80 & 1.66M \\
w/o retrospective & 91.40 & 1.63M \\
\midrule
\addlinespace[2pt]
\multicolumn{3}{c}{\textit{Expand the Vocabulary}} \\
\midrule
In-cat Expansion & 91.60 & 1.65M \\
New Category & 91.20 & 1.69M \\
\bottomrule
\end{tabular}
}
\caption{Ablation and expansion of trigger categories.}\vspace{-0.4cm}
\label{tab:vocab_ablation}
\end{threeparttable}
\end{table}

% \begin{table}[ht]
% \centering
% \small
% \begin{tabular}{l|cc}
% \hline
% \textbf{Variants} & \textbf{Pass@1}$\uparrow$ & \textbf{\#Tokens}$\downarrow$ \\
% \hline
% Vanilla & 91.40 & 2.64M \\
% REFRAIN (no-valid) & 91.20 & 1.61M \\
% \hline
% \addlinespace[1pt]
% \multicolumn{3}{c}{\textit{Shorten the Vocabulary}} \\
% \hline
% w/o self-check & 90.20 & 1.77M \\
% w/o strategy-shift & 90.20 & 1.68M \\
% w/o uncertainty & 90.80 & 1.66M \\
% w/o retrospective & 91.40 & 1.63M \\
% \hline
% \addlinespace[1pt]
% \multicolumn{3}{c}{\textit{Expand the Vocabulary}} \\
% \hline
% In-cat Expansion & 91.60 & 1.65M \\
% New Category & 91.20 & 1.69M \\
% \hline
% \end{tabular}
% \caption{Title of this table}
% \label{tab:my_label}
% \end{table}

According to Table \ref{tab:vocab_ablation}, the removal of self-check or strategy-shift triggers leads to a notable drop in Pass@1, while removing uncertainty or retrospective triggers has minor effects. Moreover, expanding the vocabulary within existing categories or adding a new category does not yield further improvements. These findings suggest that our four categories already capture the essential reflective behaviors, with self-check and strategy-shift being particularly critical, while the current vocabulary provides sufficient coverage without requiring further expansion.

% According to Table \ref{tab:vocab_ablation}, neither expanding the existing category nor adding new categories significantly changes the results. It indicates that the current four categories and their included phrases already cover the semantic functions required for triggering.
\vspace{-0.1cm}

% Under our settings, both Sentence-Transformer and SimCSE significantly reduce tokens while maintaining nearly no drop in accuracy. In contrast, TF-IDF and ROUGE-L exhibit less early stopping, resulting in a higher number of generated tokens. Furthermore, accuracy shows a notable decline under the ROUGE-L setting. We think this may be because TF-IDF primarily captures word-level overlap, while ROUGE-L primarily captures the longest common subsequence. When models repeat the same reasoning using different expressions, the actual redundancy remains unrecognized, resulting in a failure to stop thinking.

\subsection{Importance of UCB in Optimal Threshold Selection}

Although our method adaptively selects thresholds based on sliding-window UCB and demonstrates stable performance in experiments, it remains to be verified whether the UCB strategy is essential or if simpler heuristics could achieve comparable results. To this end, we compared it against three alternative early-stopping heuristics on the MATH-500 benchmark using Qwen3-8B:

\noindent \textbf{Step-based Probability Early Stop (SP-Early Stop):} Each step terminates reasoning early with a probability that increases linearly with the number of steps, where $Stop\_prob(n) = min(0.5, 2\times10^{-3}\times n)$, effective from 10th step to prevent insufficient thinking.

\noindent \textbf{Step-based Probability Early Stop with Trigger Words (SPTW-Early Stop):} Stop thinking based on probability only when trigger words are included in the generated steps. The remaining settings remain the same as SP-Early Stop.

\noindent \textbf{Randomly Select Threshold (RST):} Instead of using UCB to optimize the threshold, a threshold is randomly selected for each problem from the candidate threshold set, simulating a scenario of no exploration for optimal thresholds.

\begin{table}[ht]
\vspace{-0.3cm}
\centering
\small
\begin{threeparttable}
\setlength{\tabcolsep}{8pt}
\resizebox{0.4\textwidth}{!}{
\begin{tabular}{@{}l|cc@{}}
\toprule
\textbf{Method} & \textbf{Pass@1}$\uparrow$ & \textbf{\#Tokens}$\downarrow$ \\
\midrule
Vanilla & 91.40 & 2.64M \\
% \midrule
SP-Early Stop & 68.60 & 0.74M \\
% \midrule
SPTW-Early Stop & 83.20 & 0.91M \\
% \midrule
RST & 88.40 & 1.68M \\
\midrule
REFRAIN & 91.20 & 1.61M \\
\bottomrule
\end{tabular}
}
\caption{Comparison of REFRAIN with alternative early-stopping heuristics.}\vspace{-0.3cm}
\label{tab:compare_refrain}
\end{threeparttable}
\end{table}

Table \ref{tab:compare_refrain} demonstrate that purely probability-based stopping (SP-Early Stop) severely harms accuracy despite reducing token usage, indicating that indiscriminate truncation prematurely cuts off necessary reasoning. Incorporating trigger words (SPTW-Early Stop) partially mitigates this issue but still falls short of our method, suggesting that lexical cues alone are insufficient for robust stopping. Random threshold selection (RST) yields moderate accuracy but fails to balance efficiency and correctness. By contrast, REFRAIN consistently achieves a better trade-off between accuracy and efficiency, confirming that trigger words, semantic redundancy scorer, adaptively selecting thresholds are crucial for stable performance.

% Based on the results, we draw the following conclusions. First, purely probability-based early stopping is not feasible. Although token consumption is minimized, the accuracy drops sharply to 68.6\%. This demonstrates that blindly stopping based on probability truncates the reasoning chain prematurely, severely compromising the answer quality. Then, adding trigger words can partially mitigate this problem. Although the accuracy improves, it remains significantly lower than that of our proposed method. It shows that the 'semantic + random' mechanism still lacks robustness. Finally, using trigger words and semantic similarity can further enhance the results. However, without UCB to select the optimal threshold, this method struggles to balance accuracy and efficiency. The above findings demonstrate that semantic similarity, trigger words, and UCB are essential: they enable our proposed method to adaptively adjust thresholds across different problems, achieving an optimal accuracy-efficiency trade-off.

\subsection{Is Answer-Only Likelihood Enough?}

REFRAIN uses the average log-likelihood of the boxed final answer as the reward signal, which naturally aligns with the pre-training objective of decoder-only LLMs and avoids additional forward passes. A natural question is whether external reward models (RMs/PRMs), which can evaluate full solutions or step-wise reasoning quality, offer better alignment for early stopping. To address this, we replace the likelihood reward with two widely used alternatives and the experiments are conducted on GSM8K and MATH-500 using Qwen3-8B:

\noindent \textbf{AceMath-7B.} AceMath-7B \cite{acemath2024} is a math reward model (RM) that assigns a scalar score to the complete solution.

\noindent \textbf{Qwen2.5-MATH-PRM-7B.} Qwen2.5-MATH-PRM-7B \cite{prmlessons} is a math process reward model (PRM) that assigns a score to each step of the solution, and we take the mean as the final score.

\begin{table}[ht]
\vspace{-0.2cm}
\centering
\small
\begin{threeparttable}
\setlength{\tabcolsep}{5pt}
\resizebox{0.49\textwidth}{!}{
\begin{tabular}{@{}l|cc|cc@{}}
\toprule
\multirow{2}{*}{\textbf{Method}} 
& \multicolumn{2}{c|}{\textbf{GSM8K}} 
& \multicolumn{2}{c}{\textbf{MATH-500}} \\
\cmidrule(lr){2-3} \cmidrule(lr){4-5}
& \textbf{Pass@1}$\uparrow$ & \textbf{\#Tokens}$\downarrow$ 
& \textbf{Pass@1}$\uparrow$ & \textbf{\#Tokens}$\downarrow$ \\
\midrule
Vanilla             & 94.24 & 2.62M & 91.40 & 2.64M \\
AceMath-7B          & 94.47 & 1.71M & 90.60 & 1.70M \\
Qwen2.5-MATH-PRM-7B & 94.24 & 1.66M & 91.60 & 1.74M \\
\midrule
REFRAIN (Likelihood)             & 94.54 & 1.68M & 91.20 & 1.61M \\
\bottomrule
\end{tabular}
}
\caption{Likelihood vs. external reward models.}
\label{tab:gsm8k_math500}
\end{threeparttable}
\vspace{-0.2cm}
\end{table}

% \begin{table}[ht!]
% \centering
% \begin{tabular}{l|cc|cc}
% \hline \hline
% \multirow{2}{*}{\textbf{Method}} & \multicolumn{2}{c|}{\textbf{GSM8K}} & \multicolumn{2}{c}{\textbf{MATH-500}} \\
% \cline{2-5}
% & \textbf{Pass@1}$\uparrow$ & \textbf{\#Tokens}$\downarrow$ & \textbf{Pass@1}$\uparrow$ & \textbf{\#Tokens}$\downarrow$ \\
% \hline
% Vanilla & 94.24 & 2.62M & 91.40 & 2.64M \\
% \hline
% AceMath-7B & 94.47 & 1.71M & 90.60 & 1.70M \\
% \hline
% Qwen2.5-MATH-PRM-7B & 94.24 & 1.66M & 91.60 & 1.74M \\
% \hline
% REFRAIN & 94.54 & 1.68M & 91.20 & 1.61M \\
% \hline \hline
% \end{tabular}
% \caption{title}
% \label{tab:my_label}
% \end{table}

% As summarized in Table 8, the PRM delivers accuracy on par with answer-only likelihood, while the scalar RM underperforms likelihood on MATH-500. Crucially, both RM and PRM introduce an additional forward pass and memory overhead beyond generation, whereas the likelihood-based objective incurs no extra compute. Consequently, the answer-only likelihood achieves the best accuracy–efficiency trade-off under our early-stopping framework and is a practical default for CoT adaptive stopping on Qwen3-8B across GSM8K and MATH-500.

As summarized in Table \ref{tab:gsm8k_math500}, the PRM provides a Pass@1 equal to the answer-only likelihood, while the scalar RM underperforms the likelihood. However, both RM and PRM require an additional forward pass, increasing computational cost, whereas the likelihood-based objective incurs no extra compute. Therefore, the answer-only likelihood achieves the best accuracy-efficiency trade-off under our early-stopping framework and is a practical default for CoT adaptive stopping.

\subsection{Test-time Scaling with Budgeted Thinking}

We treat budgeted thinking as test-time scaling that dynamically adjusts reasoning length. Concretely, for a given instance, we limit the cumulative number of thinking tokens. Once the limit is reached, the model is forced to stop the reasoning chain and generate the final answer. We sweep the budget from small to large while keeping all other decoding settings unchanged, and evaluate on Qwen3-8B across the four benchmarks.

The experimental results are shown in Figure \ref{fig:scaling} and numerical values are provided in Table \ref{tab:scaling}. For the relatively simple datasets GSM8K and CSQA, optimal results can be approximated using fewer reasoning tokens. Our adaptive method rapidly detects and stops reasoning, thus achieving higher or comparable accuracy with fewer tokens. MATH-500 and GPQA-Diamond contain more problems that require longer reasoning to perform well. In contrast, our method can allocate more reasoning budget to challenging problems, thereby achieving a better accuracy-cost trade-off.

Conceptually, test-time scaling usually samples more chains or extends chain length. We show that learning when to stop within a single chain is similarly effective: it balances accuracy and tokens without extra passes, reallocating computation per instance. Uniform fixed budgets over-allocate easy instances and under-allocate hard ones. In contrast, adaptive stopping strategies continuously adjust to achieve the optimal trade-off.

% From the results, increased budgets lead to higher accuracy, consistent with findings from test-time scaling research. However, setting a consistent budget for all benchmarks may lead to over-reasoning on easy problems while under-reasoning on difficult ones. Therefore, a consistent budget cannot achieve the optimal accuracy-cost tradeoff. 

% Test-time scaling typically achieves accuracy gains by increasing the number of reasoning chains or the number of reasoning steps. Our results demonstrate that learning a strategy for when to stop within a single reasoning chain is also a type of test-time scaling. It does not involve more computation, but rather smarter use of the same or fewer computational resources.

\subsection{Bandit Variants: \texorpdfstring{$\epsilon$}{epsilon}-greedy MAB vs. SW-UCB}

Our adaptive threshold method requires online selection of the current question's threshold from a set of similarity thresholds $\mathcal{T}$, thereby balancing `token savings' with `generating correct answers.' Because task difficulty and model state vary over time, the optimal threshold is non-stationary. Therefore, we compare SW-UCB \cite{garivier2008upper} with another classic multi-arm bandit ($\epsilon$-greedy MAB) strategy \cite{sutton1998reinforcement}. Specifically, $\epsilon$-greedy explores any arm $t \in \mathcal{T}$ with probability $\epsilon$ or selects the arm $t$ with the highest average reward $\bar{R}_t$ with probability 1-$\epsilon$:

\begin{equation}
t = \arg \max_{t \in \mathcal{T}} \bar{R}_t, 
\quad \bar{R}_t = \frac{\sum R_t}{n_t}.
\end{equation}

The reward definition follows Eq. \ref{eq:reward}.

% \begin{table}[ht]
% \centering
% \small
% \begin{threeparttable}
% \setlength{\tabcolsep}{6pt}
% \begin{tabular}{@{}l|cc|cc|cc|cc@{}}
% \toprule
% \multirow{2}{*}{\textbf{Method}} 
% & \multicolumn{2}{c|}{\textbf{GSM8K}} 
% & \multicolumn{2}{c|}{\textbf{MATH-500}}
% & \multicolumn{2}{c|}{\textbf{CSQA}} 
% & \multicolumn{2}{c}{\textbf{GPQA-Diamond}} \\
% \cmidrule(lr){2-3} \cmidrule(lr){4-5} \cmidrule(lr){6-7} \cmidrule(lr){8-9}
% & \textbf{Pass@1}$\uparrow$ & \textbf{\#Tokens}$\downarrow$ 
% & \textbf{Pass@1}$\uparrow$ & \textbf{\#Tokens}$\downarrow$
% & \textbf{Pass@1}$\uparrow$ & \textbf{\#Tokens}$\downarrow$
% & \textbf{Pass@1}$\uparrow$ & \textbf{\#Tokens}$\downarrow$\\
% \midrule
% Vanilla             & 94.24 & 2.62M & 91.40 & 2.64M \\
% AceMath-7B          & 94.47 & 1.71M & 90.60 & 1.70M \\
% Qwen2.5-MATH-PRM-7B & 94.24 & 1.66M & 91.60 & 1.74M \\
% \midrule
% REFRAIN (Likelihood)             & 94.54 & 1.68M & 91.20 & 1.61M \\
% \bottomrule
% \end{tabular}
% \caption{Likelihood vs. external reward models.}
% \label{tab:gsm8k_math500}
% \end{threeparttable}
% \end{table}

\begin{table}[ht!]
\vspace{-0.2cm}
\centering
\small
\begin{threeparttable}
\setlength{\tabcolsep}{6pt}
\resizebox{0.4\textwidth}{!}{
\begin{tabular}{@{}l|cc|cc@{}}
\toprule
\multirow{2}{*}{\textbf{Method}} 
& \multicolumn{2}{c|}{\textbf{GSM8K}} 
& \multicolumn{2}{c}{\textbf{MATH-500}} \\
\cmidrule(lr){2-3} \cmidrule(lr){4-5}
& \textbf{Pass@1}$\uparrow$ & \textbf{\#Tokens}$\downarrow$ 
& \textbf{Pass@1}$\uparrow$ & \textbf{\#Tokens}$\downarrow$ \\
\midrule
MAB     & 94.62 & 1.70M & 91.20 & 1.72M \\
SW-UCB  & 94.54 & 1.68M & 91.20 & 1.61M \\
\bottomrule
\end{tabular}}

\vspace{0.2cm}

\resizebox{0.4\textwidth}{!}{
\begin{tabular}{@{}l|cc|cc@{}}
\toprule
\multirow{2}{*}{\textbf{Method}} 
& \multicolumn{2}{c|}{\textbf{CSQA}} 
& \multicolumn{2}{c}{\textbf{GPQA-Diamond}} \\
\cmidrule(lr){2-3} \cmidrule(lr){4-5}
& \textbf{Pass@1}$\uparrow$ & \textbf{\#Tokens}$\downarrow$ 
& \textbf{Pass@1}$\uparrow$ & \textbf{\#Tokens}$\downarrow$ \\
\midrule
MAB     & 82.47 & 0.79M & 55.05 & 1.39M \\
SW-UCB  & 84.03 & 0.76M & 60.10 & 1.42M \\
\bottomrule
\end{tabular}}
\end{threeparttable}
\vspace{-0.2cm}
\caption{$\epsilon$-greedy MAB vs. sliding-window UCB for adaptive threshold selection.}
\label{tab:adaptive threshold selection}
\end{table}

Table \ref{tab:adaptive threshold selection} shows that both methods perform comparably on GSM8K and MATH-500, with SW-UCB generating slightly fewer tokens. On CSQA and GPQA-Diamond, where the question semantics fluctuate more, SW-UCB attains higher accuracy at similar or lower token budgets. These results indicate that SW-UCB better adapts to non-stationary reward dynamics, providing more consistent thresholding behavior. 
Overall, SW-UCB is a more robust choice: it conserves tokens in relatively stable settings and improves accuracy in volatile ones.

\section{Conclusion}

We proposed REFRAIN, a training-free framework that mitigates overthinking by detecting reflective redundancy and adaptively tuning stop thresholds via a sliding-window UCB controller.
Across four benchmarks and two model families, REFRAIN reduces token usage by 20–55\% while preserving or improving accuracy, establishing when-to-stop as a practical axis of test-time scaling for efficient and reliable reasoning.

% We introduced a dynamic framework to mitigate overthinking in reasoning LLMs by detecting reflective redundancy and adaptively tuning stop thresholds with a sliding-window UCB bandit. Experiments on four benchmarks and two LLMs show that our method consistently reduces token usage by 30–60\% while maintaining or improving accuracy. Ablation studies confirm the importance of semantic redundancy detection and the necessity of UCB-based thresholding. Overall, our results highlight `when to stop' as a practical axis of test-time scaling, offering an efficient and reliable solution for reasoning LLMs.

% \clearpage
% \newpage
\section*{Limitations}

REFRAIN relies on observable step-by-step reasoning during inference, which means the model must expose or stream intermediate reasoning traces so the system can determine when to stop. This assumption is common across most test-time methods for mitigating overthinking or adaptively allocating compute (e.g., stopping rules, reflection loops, or debate-style reasoning). For closed-source APIs that only return final answers without providing step-by-step control, REFRAIN, like other similar methods, cannot intervene during the generation process. Nevertheless, our framework remains applicable to any setting where partial reasoning signals or token-level outputs are available, including open-source and step-streaming models.
\bibliography{custom}
\clearpage
\newpage

\appendix

% ===================== Appendix A =====================
\section*{Appendix A \quad Reflection Trigger Vocabulary}
\label{app:reflection_lexicon}

This appendix documents the reflection trigger vocabulary used by the
two-stage stop discriminator (Sec.~\ref{subsec:discriminator}).
We group phrases into four categories that correspond to self-check, strategy shifts, expressed uncertainty, and retrospective revisions.
The following lists are verbatim and were used as is in all experiments unless otherwise specified.

\begin{table*}[ht]
\centering
\footnotesize
\setlength{\tabcolsep}{6pt}
\begin{tabular}{p{3.1cm} p{10.8cm}}
\hline
\textbf{Category} & \textbf{Trigger phrases} \\
\hline
Self Check (\(V_{\text{check}}\)) &
\texttt{wait}; \texttt{let me check}; \texttt{hold on}; \texttt{have made a mistake}; \texttt{\underline{let me double check}}; \texttt{\underline{wait a moment}}; \texttt{\underline{is that correct}}; \texttt{\underline{let me re-read}} \\
\addlinespace[2pt]
Strategy Shift (\(V_{\text{shift}}\)) &
\texttt{alternatively}; \texttt{let me try}; \texttt{think of it as}; \texttt{let me consider}; \texttt{\underline{what if we try}}; \texttt{\underline{let's think from a different angle}}; \texttt{\underline{an alternative method would be}}; \texttt{\underline{instead of doing that}} \\
\addlinespace[2pt]
Uncertainty (\(V_{\text{uncert}}\)) &
\texttt{not sure}; \texttt{looks like}; \texttt{that seems}; \texttt{hmm}; \texttt{perhaps}; \texttt{maybe i}; \texttt{\underline{i'm not certain}}; \texttt{\underline{it seems}}; \texttt{\underline{i suspect}}; \texttt{\underline{my guess is}} \\
\addlinespace[2pt]
Retrospective (\(V_{\text{retro}}\)) &
\texttt{earlier we saw}; \texttt{from before}; \texttt{so now we have}; \texttt{recall that}; \texttt{let me go back}; \texttt{\underline{as we established previously}}; \texttt{\underline{based on our previous result}}; \texttt{\underline{remember that we found}}; \texttt{\underline{the value from step}} \\
\addlinespace[2pt]
\textbf{New Category} & \textbf{simplify this problem}; \textbf{the core of the problem is}; \textbf{this is equivalent to}; \textbf{this is equal to}; \textbf{the key insight here is}; \textbf{break this down}; \textbf{the overall plan is to}; \textbf{the plan is to} \\
\hline
\end{tabular}
\caption{Reflection trigger vocabulary \(V = V_{\text{check}}\cup V_{\text{shift}}\cup V_{\text{uncert}}\cup V_{\text{retro}}\). Besides, we use \underline{underline} to indicate the In-cat Expansion and \textbf{bold} to indicate the New Category in Section \ref{sec:triggers}.}
\label{tab:reflection-lexicon}
\end{table*}

\begin{table*}[ht!]
\centering
\small
\begin{threeparttable}
\setlength{\tabcolsep}{6pt}
\begin{tabular}{@{}l|cc|cc|cc|cc@{}}
\toprule
\multirow{2}{*}{\textbf{Method}} 
& \multicolumn{2}{c|}{\textbf{GSM8K}} 
& \multicolumn{2}{c|}{\textbf{MATH-500}} 
& \multicolumn{2}{c|}{\textbf{CSQA}} 
& \multicolumn{2}{c}{\textbf{GPQA-Diamond}} \\
\cmidrule(lr){2-3} \cmidrule(lr){4-5} \cmidrule(lr){6-7} \cmidrule(lr){8-9}
& \textbf{Pass@1}$\uparrow$ & \textbf{\#Tokens}$\downarrow$ 
& \textbf{Pass@1}$\uparrow$ & \textbf{\#Tokens}$\downarrow$ 
& \textbf{Pass@1}$\uparrow$ & \textbf{\#Tokens}$\downarrow$
& \textbf{Pass@1}$\uparrow$ & \textbf{\#Tokens}$\downarrow$ \\
\midrule
Vanilla ($P_0$)     & 94.24 & 2.62M & 91.40 & 2.64M & 83.13 & 1.66M & 53.54 & 1.81M \\
REFRAIN ($P_0$)  & 94.54 & 1.68M & 91.20 & 1.61M & 84.03 & 0.76M & 60.10 & 1.42M \\
\midrule
Vanilla ($P_1$)     & 94.69 & 2.84M & 91.20 & 2.28M & 84.77 & 1.83M & 61.11 & 1.88M\\
REFRAIN ($P_1$)  & 94.77 & 1.89M & 91.00 & 1.69M & 82.56 & 0.79M & 61.11 & 1.35M \\
\midrule
Vanilla ($P_2$)     & 94.92 & 2.45M & 92.00 & 2.52M & 83.21 & 1.80M & 59.60 & 1.91M \\
REFRAIN ($P_2$)  & 94.92 & 1.58M & 92.20 & 1.68M & 82.80 & 0.79M & 59.09 & 1.42M \\
\bottomrule
\end{tabular}
\end{threeparttable}

\caption{REFRAIN vs. Vanilla under prompt variants $P_0$-$P_2$.}
\label{tab:prompt_robustness}
\end{table*}

% \paragraph{Minimal reproducibility notes.}
% Unless otherwise specified, we match triggers \emph{case-insensitively} over each decoded step \(s_n\) using substring search (punctuation ignored), and we do not stem/lemmatize.
% Multiple matches within the same step are treated as a single reflection flag.
% To reduce early false positives, the reflection condition is combined with the provisional-answer cue \(c\) as described in Sec.~\ref{subsec:discriminator}.

\section*{Appendix B \quad Results of Test-time Scaling with Budgeted Thinking}
\label{app:scaling}

The results are shown in Table \ref{tab:scaling}.

\begin{table}[ht]
\centering
\small
\begin{threeparttable}
\setlength{\tabcolsep}{6pt}
\begin{tabular}{@{}l|cc@{}}
\toprule
\textbf{GSM8K} & \textbf{Pass@1}$\uparrow$ & \textbf{\#Tokens}$\downarrow$ \\
\midrule
        Vanilla & 94.31 & 2.36M \\
        % \hline
        Budget (512 tokens) & 76.12 & 0.88M \\
        % \hline
        Budget (1024 tokens) & 89.76 & 1.40M \\
        % \hline
        Budget (1536 tokens) & 92.27 & 1.81M \\
        \midrule
        REFRAIN & 94.54 & 1.68M \\
\bottomrule
\end{tabular}

\hspace{0.01\textwidth}

\begin{tabular}{@{}l|cc@{}}
\toprule
\textbf{MATH-500} & \textbf{Pass@1}$\uparrow$ & \textbf{\#Tokens}$\downarrow$ \\
\midrule
        Vanilla & 91.40 & 2.64M \\
        % \hline
        Budget (1024 tokens) & 69.20 & 0.60M \\
        % \hline
        Budget (2048 tokens) & 80.80 & 1.05M \\
        % \hline
        Budget (3072 tokens) & 84.20 & 1.41M \\
        \midrule
        REFRAIN & 91.20 & 1.61M \\
\bottomrule
\end{tabular}

\hspace{0.01\textwidth}

\begin{tabular}{@{}l|cc@{}}
\toprule
\textbf{CSQA} & \textbf{Pass@1}$\uparrow$ & \textbf{\#Tokens}$\downarrow$ \\
\midrule
        Vanilla & 83.13 & 1.66M \\
        % \hline
        Budget (256 tokens) & 80.02 & 0.38M \\
        % \hline
        Budget (512 tokens) & 82.64 & 0.76M \\
        % \hline
        Budget (768 tokens) & 82.80 & 1.03M \\
        \midrule
        REFRAIN & 84.03 & 0.76M \\
\bottomrule
\end{tabular}

\hspace{0.01\textwidth}

\begin{tabular}{@{}l|cc@{}}
\toprule
\textbf{GPQA-Diamond} & \textbf{Pass@1}$\uparrow$ & \textbf{\#Tokens}$\downarrow$ \\
\midrule
        Vanilla & 53.54 & 1.81M \\
        % \hline
        Budget (2048 tokens) & 40.91 & 0.42M \\
        % \hline
        Budget (4096 tokens) & 51.52 & 0.83M \\
        % \hline
        Budget (8192 tokens) & 54.04 & 1.38M \\
        \midrule
        REFRAIN & 60.10 & 1.42M \\
\bottomrule
\end{tabular}
\caption{Test-time scaling with fixed thinking budgets vs. adaptive stopping.}
\label{tab:scaling}
\end{threeparttable}
\end{table}

\section*{Appendix C \quad Alternatives for the Redundancy Scorer}
% \vspace{-0.1cm}
REFRAIN's early stopping strategy relies on a semantic redundancy scorer to measure similarity between the current step and historical steps. To test robustness, we compared the default sentence-transformer encoder (SBERT) with alternative similarity functions, including another embedding-based model (SimCSE) and two sequence overlap metrics (TF-IDF, ROUGE-L). All experiments were conducted using the Qwen3-8B model on the MATH-500 benchmark.

% Our early stopping strategy relies on a `semantic redundancy scorer' to measure similarity between the current step and historical steps. Therefore, the choice of similarity function may directly impact the trade-off between trigger frequency and final accuracy/cost. To evaluate the robustness and substitutability of our approach, we compared semantic sentence vectors (Sentence Transformer, SimCSE) with lexical/sequence overlap metrics (TF-IDF, ROUGE-L). We replace the encoder $f(\cdot)$ in the semantic redundancy scorer with three mentioned common similarity modules SimCSE, TF-IDF, and ROUGE-L. We also keep other components unchanged. 

\begin{table}[ht]
\vspace{-0.3cm}
\centering
\small
\begin{threeparttable}
\setlength{\tabcolsep}{8pt}
\resizebox{0.4\textwidth}{!}{
\begin{tabular}{@{}l|cc@{}}
\toprule
\textbf{Method} & \textbf{Pass@1}$\uparrow$ & \textbf{\#Tokens}$\downarrow$ \\
\midrule
Vanilla & 91.40 & 2.64M \\
\midrule
\addlinespace[2pt]
\multicolumn{3}{c}{\textit{Embedding-based Similarity Metrics}} \\
\midrule
% \cmidrule(lr){1-3}
SBERT (default) & 91.20 & 1.61M \\
SimCSE           & 91.40 & 1.67M \\
\midrule
\addlinespace[2pt]
\multicolumn{3}{c}{\textit{Sequence Overlap Metrics}} \\
\midrule
TF-IDF  & 91.20 & 1.91M \\
Rouge-L & 90.40 & 1.92M \\
\bottomrule
\end{tabular}
}
\caption{Comparison of alternative redundancy scorers}\vspace{-0.3cm}
\label{tab:metrics}
\end{threeparttable}
\end{table}

The results in Table \ref{tab:metrics} show that both SBERT and SimCSE achieve substantial token reduction with minimal accuracy loss, indicating that embedding-based semantic representations are effective for detecting redundancy. In contrast, TF-IDF and ROUGE-L yield higher token usage, suggesting that surface-level overlap metrics fail to capture paraphrased or semantically redundant reasoning steps. These findings confirm that embedding-based scorers best capture reflective redundancy in CoT reasoning. 

\section*{Appendix D \quad Prompt Robustness}

While all main experiments use a single instruction template $P_0$, reasoning behavior can be sensitive to small prompt wording changes. To ensure that our conclusions are not artifacts of a particular wording choice, we evaluate prompt robustness by re-running the same method and configurations under two paraphrased prompt templates $P_1$ and $P_2$. We evaluate on four benchmarks using Qwen3-8B. The three prompts are:

\begin{itemize}
\item $P_0:$ \{question\}\textbackslash nPlease answer step by step. End your response with: Final Answer: \textbackslash boxed\{your final answer here\}. Make sure to wrap your final answer in \textbackslash boxed\{\}.
\item $P_1:$ You are a helpful AI Assistant, designed to provided well-reasoned and detailed responses. You FIRST think about the reasoning process step by step and then provide the user with the answer. \textbackslash nQuestion: \{question\}\textbackslash nPlease enclose your final answer in the box: Final Answer: \textbackslash boxed\{Your Answer\}.
\item $P_2:$ Please solve the following question. Question: \{question\}\textbackslash n\textbackslash nAfter reasoning step by step, conclude with the final answer in the format: Final Answer: \textbackslash boxed\{Your Answer\}.
\end{itemize}

% This analysis addresses two questions: (i) whether our accuracy gains persist across prompt phrasings, and (ii) whether our token-savings remain stable when the instruction wording changes.

% Existing research indicates that large models are sensitive to prompt format: the wording of instructions, whether step-by-step reasoning is explicitly requested, and many other factors can change the model's output. Our goal is not to find an optimal prompt, but to verify whether this method consistently reduces generation costs without sacrificing accuracy when the decoding configuration remains unchanged and only the prompt text is changed. If the conclusion stands, it indicates that our method identifies overthinking thinking signals independent of specific wording, demonstrating transferability across prompts.
According to Table \ref{tab:prompt_robustness}, REFRAIN maintains vanilla-level accuracy --- sometimes modestly better --- while consistently using far fewer thinking tokens across three paraphrased templates ($P_0$-$P_2$) on four benchmarks. In addition, changing prompts can cause significant Pass@1 and token drift in vanilla models, such as Pass@1 ranging from 53.54 to 61.11 on GPQA-Diamond and \#Tokens ranging from 2.28M to 2.64M on MATH-500. REFRAIN effectively stabilizes model behavior under prompt perturbations, yielding stable cost with no catastrophic accuracy drops. Overall, these results indicate that REFRAIN is robust to reasonable prompt rewrites and reduces the need for prompt-specific tuning during deployment.

\section*{Appendix E \quad Implementation Details}

For Qwen3-8B, we enable the official thinking mode and adopt the recommended decoding configuration: Temperature = 0.6, Top-p = 0.95, and Top-k = 20.

For gpt-oss-20B, we use bfloat16 precision and apply the recommended decoding configuration: Temperature = 1.0, Top-p = 1.0, and Top-k = 50.

We deploy the models on NVIDIA A100 GPUs using PyTorch and HuggingFace Transformers \cite{wolf2019huggingface}. The stopping threshold $\tau$ is discretized over [0.60, 0.80] with a step size of 0.05. This range empirically captures the transition between conservative and aggressive stopping behavior. We set the maximum generation length to 16,384 tokens unless otherwise specified. The random seed is set to 42.

\section*{Appendix F \quad Case Study}
\label{app:case_study}

To understand how our UCB-based early-stopping mechanism shapes model reasoning, we examined representative examples across datasets. 
In these domains, early stopping acts as a selective gate on the decoding trajectory: it halts continuation once confidence peaks, preserving early coherent reasoning while discarding later, low-confidence expansions. 
This mechanism can prevent semantic or numeric drift—but it can also truncate necessary self-corrections. 
The four cases below illustrate both sides of this trade-off within a unified narrative.

When early stop succeeds, it prevents \textbf{over-elaboration}, \textbf{re-computation}, or \textbf{format degradation}---errors that typically appear after the model has already reached a correct intermediate conclusion. 
When it fails, it interrupts \textbf{self-repair}: the model has not yet completed its reasoning, but the confidence temporarily spikes on an incorrect intermediate token, causing premature termination.

Across all examples, early stopping behaves like a \textit{confidence filter}---it favors precision at the cost of completeness. 
In commonsense and physical reasoning (CSQA, GPQA), it prunes away irrelevant elaboration or unstable numerical recomputation. 
In multi-step quantitative reasoning (MATH, GSM8K-type), it can freeze an intermediate value before the chain converges to the correct expression. 
Overall, early stopping helps when the first coherent burst already contains the solution essence, but harms when correctness emerges only after longer deliberation.

\definecolor{darkgreen}{rgb}{0.0,0.45,0.0}
\definecolor{darkred}{rgb}{0.8,0.0,0.0} % For clear contrast
\begin{table*}[ht!]
\centering
\small
\renewcommand{\arraystretch}{1.25} % 增加行高避免拥挤
\setlength{\tabcolsep}{6pt}

% 定义一个方便的顶对齐 parbox 单元格命令
\newcommand{\TopCell}[1]{\parbox[t]{\hsize}{\RaggedRight #1}}

\begin{tabularx}{\textwidth}{%
    >{\raggedright\arraybackslash}p{1.6cm}  % Dataset
    >{\raggedright\arraybackslash}X         % Question (flexible)
    >{\centering\arraybackslash}p{1.0cm}    % Gold
    >{\raggedright\arraybackslash}X         % Baseline
    >{\raggedright\arraybackslash}X         % Ours
}
\toprule
\textbf{Dataset} & \textbf{Question} & \textbf{Gold} & \textbf{Baseline} & \textbf{Ours} \\
\midrule

% CSQA row
CSQA &
\TopCell{\textbf{Where do adults use glue sticks?}\\
(A) classroom \quad (B) desk drawer\\
(C) at school \quad (D) office \quad (E) kitchen drawer} &
\TopCell{D} &
\TopCell{\dots reasons that glue sticks are for sticking papers, then shifts to ``classroom'' context; overthinks into child-use scenario.\\
\textcolor{darkred}{\textbf{Final Answer: A (classroom)}}} &
\TopCell{\dots identifies adults use them in professional settings, stops at ``office'' without drift.\\
\textcolor{darkgreen}{\textbf{Final Answer: D (office)}}} \\

\addlinespace[6pt] % 加一点竖直间距，使视觉更舒适

% GPQA row
GPQA &
\TopCell{\textbf{Observations of a quasar across the electromagnetic spectrum are being carried out.}\\
Focusing on the near infrared and optical spectrum, there is a peak at wavelength of about 790 nm, and at shorter wavelengths $<$ 790 nm the flux drops significantly.\\
If we live in a flat $\Lambda$CDM universe ($H_0=70$ km s$^{-1}$ Mpc$^{-1}$, $\Omega_m=0.3$, $\Omega_\Lambda=0.7$), what is the most likely comoving distance (for $a=1$) of this quasar from the Earth?\\
(A) 8 Gpc \quad (B) 7 Gpc \quad (C) 6 Gpc \quad (D) 9 Gpc} &
\TopCell{A} &
\TopCell{\dots computes $\sim$6\,Gpc correctly at first, then re-derives equations with slightly altered constants, inflating to \textcolor{darkred}{\textbf{9\,Gpc (D)}}.\\
Numeric hallucination from over-expansion.} &
\TopCell{\dots halts right after consistent calculation, preserving \textcolor{darkgreen}{\textbf{6\,Gpc (A)}} before re-computation noise.} \\

\addlinespace[6pt]

% MATH row 1
MATH &
\TopCell{\textbf{Find the product $6_8\times7_8$ and express the result in base 8.}} &
\TopCell{$52_8$} &
\TopCell{\dots correctly gets $42_{10}$, but continues verbose explanation, drops subscript and outputs \textcolor{darkred}{\textbf{52}}.\\
Formatting error after over-elaboration.} &
\TopCell{\dots stops immediately after producing boxed \textcolor{darkgreen}{\textbf{$52_8$}}, avoiding format decay.} \\

\addlinespace[6pt]

% MATH row 2
MATH &
\TopCell{\textbf{In a triangle with $AB=BC$ and $BC\parallel$ a line through $A$, find $\angle x$.}} &
\TopCell{$28^\circ$} &
\TopCell{\dots initially gets $60^\circ$, continues reasoning, self-corrects to \textcolor{darkgreen}{\textbf{$28^\circ$}} after applying geometric constraint.} &
\TopCell{\dots confidence peaks early, stops at intermediate \textcolor{darkred}{\textbf{$62^\circ$}}, missing late correction.} \\

\bottomrule
\end{tabularx}

\caption{Representative examples showing how early stopping prevents or causes specific reasoning errors. Green = correct; Red = incorrect. CSQA and GPQA questions and options are taken verbatim from the original datasets.}
\label{tab:case_study_full}
\end{table*}

\definecolor{darkgreen}{rgb}{0.0,0.45,0.0}

\begin{table*}[t]
\centering
\small
\renewcommand{\arraystretch}{1.25}
\setlength{\tabcolsep}{5pt}

% 顶对齐单元格
\newcommand{\TopCell}[1]{\parbox[t]{\hsize}{\raggedright #1}} % Changed \RaggedRight to \raggedright and simplified

\begin{tabularx}{\textwidth}{%
    >{\raggedright\arraybackslash}p{1.6cm}  % Dataset
    >{\raggedright\arraybackslash}X         % Question (flexible)
    >{\centering\arraybackslash}p{1.0cm}    % Gold
    >{\raggedright\arraybackslash}X         % Baseline
    >{\raggedright\arraybackslash}X         % Ours
}
\toprule
\textbf{Dataset} & \textbf{Question} & \textbf{Gold} &
\textbf{Baseline} & \textbf{Ours} \\
\midrule

% --- CSQA ---
CSQA &
\TopCell{\textbf{Where do adults use glue sticks?}\\
(A) classroom \quad (B) desk drawer\\
(C) at school \quad (D) office \quad (E) kitchen drawer} &
\TopCell{D} &
\TopCell{\texttt{<think>} \dots “Glue sticks are often used in schools.” \dots “teachers (adults) use glue sticks in a classroom.” \dots \textcolor{red}{“Final Answer: A (classroom)”} \dots \texttt{</think>}} &
\TopCell{\texttt{<think>} \dots “Adults would more typically use glue sticks in an office for attaching papers.” \dots “At school is too general.” \dots \textcolor{darkgreen}{“Final Answer: D (office)”} \dots \texttt{</think>}} \\

\addlinespace[6pt]

% --- GPQA ---
GPQA &
\TopCell{\textbf{Observations of a quasar across the electromagnetic spectrum are being carried out.}\\
Focusing on the near infrared and optical spectrum, there is a peak at wavelength of about 790 nm, and at shorter wavelengths $<$ 790 nm the flux drops significantly.\\
If we live in a flat $\Lambda$CDM universe ($H_0=70$ km s$^{-1}$ Mpc$^{-1}$, $\Omega_m=0.3$, $\Omega_\Lambda=0.7$), what is the most likely comoving distance (for $a=1$) of this quasar from the Earth?\\
(A) 8 Gpc \quad (B) 7 Gpc \quad (C) 6 Gpc \quad (D) 9 Gpc} &
\TopCell{A} &
\TopCell{\texttt{<think>} \dots “$\lambda_{\mathrm{obs}}/\lambda_{\mathrm{rest}} \approx 790/400 \Rightarrow z \approx 1$.” \dots “comoving distance $\approx 6$ Gpc.” \dots then \textcolor{red}{“maybe $z=2 \rightarrow 9$ Gpc.”} \dots \textcolor{red}{“Final Answer: D (9 Gpc)”} \dots \texttt{</think>}} &
\TopCell{\texttt{<think>} \dots “At $z \approx 1$, $d_c \approx 6$ Gpc under $\Lambda$CDM.” \dots “consistent with given parameters.” \dots \textcolor{darkgreen}{“Final Answer: A (6 Gpc)”} \dots \texttt{</think>}} \\

\addlinespace[6pt]

% --- MATH 1 ---
MATH &
\TopCell{\textbf{Find the product $6_8\times7_8$ and express the result in base 8.}} &
\TopCell{$52_8$} &
\TopCell{\texttt{<think>} \dots “$6\times7=42$.” \dots “Convert 42 to base $8 \rightarrow 52_8$.” \dots then continues \dots \textcolor{red}{“drops subscript $\rightarrow$ Final Answer: 52.”} \dots \texttt{</think>}} &
\TopCell{\texttt{<think>} \dots “$6\times7=42 \rightarrow 52_8$.” \dots \textcolor{darkgreen}{“Stops right after boxed $52_8$.”} \dots \texttt{</think>}} \\

\addlinespace[6pt]

% --- MATH 2 ---
MATH &
\TopCell{\textbf{In a triangle with $AB=BC$ and $BC\parallel$ a line through $A$, find $\angle x$.}} &
\TopCell{$28^\circ$} &
\TopCell{\texttt{<think>} \dots “Parallel lines $\rightarrow$ alternate interior angles equal.” \dots \textcolor{darkgreen}{“$x=28^\circ$.”} \dots \texttt{</think>}} &
\TopCell{\texttt{<think>} \dots “maybe supplementary, $180-118=62^\circ$.” \dots \textcolor{red}{“Final Answer: $62^\circ$.”} \dots \texttt{</think>}} \\

\bottomrule
\end{tabularx}

\caption{Excerpted raw model responses showing only critical reasoning segments (dark green = correct, red = incorrect). Non-essential text is omitted as “…” for brevity.}
\label{tab:case_key_segments_fixed}
\end{table*}

\noindent
In the first case presented in Figure~\ref{tab:case_study_full}, early stopping prevented \textit{semantic drift}: the baseline continued elaborating until it replaced ``office'' with the more frequent but irrelevant ``classroom.'' 
In the second, it prevented \textit{numeric hallucination}: the baseline re-integrated equations and inflated the distance to 9\,Gpc, whereas early stop froze the stable 6\,Gpc block. 
In the third, it avoided \textit{format degradation}: stopping right after the correct $\boxed{52_8}$ preserved syntax that the baseline later corrupted to plain ``52.'' 
In the fourth, however, the same mechanism caused a \textit{truncation error}: the model had not yet applied the geometric constraint that halves the angle, so early stop froze an intermediate $62^\circ$ instead of the correct $28^\circ$. 

Together these examples reveal that early stopping serves as a precision-biased filter: 
it trims the low-probability tail of reasoning---removing noise and overthinking---but can also cut off genuine late corrections. 
In practice, combining early stop with a minimum reasoning length or explicit ``Final Answer'' check mitigates this risk while retaining its benefits.

% \section{Example Appendix}
% \label{sec:appendix}

% This is an appendix.

\end{document}